\documentclass[10pt,twocolumn,letterpaper]{article}

\usepackage[pagenumbers]{cvpr} 

\usepackage{multirow}
\usepackage{xcolor}

\definecolor{cvprblue}{rgb}{0.21,0.49,0.74}
\usepackage[pagebackref,breaklinks,colorlinks,allcolors=cvprblue]{hyperref}

\def\paperID{MORSE-9} 
\def\confName{CVPR}
\def\confYear{2025}

\title{Foundation Models for Remote Sensing: An Analysis of MLLMs for Object Localization}

\author{
Darryl Hannan, John Cooper, Dylan White, Timothy Doster, Henry Kvinge, and Yijing Watkins\\
Pacific Northwest National Laboratory\\
Seattle, WA\\
{\tt\small darryl.hannan, john.cooper, dylan.white,}\\
{\tt\small timothy.doster, henry.kvinge, yijing.watkins @pnnl.gov}
}

\begin{document}
\maketitle
\begin{abstract}
Multimodal large language models (MLLMs) have altered the landscape of computer vision, obtaining impressive results across a wide range of tasks, especially in zero-shot settings. Unfortunately, their strong performance does not always transfer to out-of-distribution domains, such as earth observation (EO) imagery. Prior work has demonstrated that MLLMs excel at some EO tasks, such as image captioning and scene understanding, while failing at tasks that require more fine-grained spatial reasoning, such as object localization. However, MLLMs are advancing rapidly and insights quickly become out-dated. In this work, we analyze more recent MLLMs that have been explicitly trained to include fine-grained spatial reasoning capabilities, benchmarking them on EO object localization tasks. We demonstrate that these models are performant in certain settings, making them well suited for zero-shot scenarios. Additionally, we provide a detailed discussion focused on prompt selection, ground sample distance (GSD) optimization, and analyzing failure cases. We hope that this work will prove valuable as others evaluate whether an MLLM is well suited for a given EO localization task and how to optimize it.
\end{abstract}

\section{Introduction}
A primary drawback of deep learning is its reliance on large annotated datasets. This can be especially challenging when training a model on an earth observation (EO) task, where annotated data is costly to acquire and large, publicly available datasets are uncommon. Multimodal large language models (MLLMs) offer a potential solution to this problem, as they are strong, general-purpose models that achieve impressive performance across a variety of tasks in zero-shot settings \citep{dubey2024llama,deitke2024molmo,qwen2.5-VL,Qwen2VL}. However, this performance does not always transfer to EO tasks due to the domain gap between overhead imagery and the web images that are used for most MLLM training pipelines \citep{zhang2024vleobench,hu2023rsgpt}. While some MLLMs have been fine-tuned specifically on EO data \citep{kuckreja2024geochat,irvin2024teochat}, there is simply not enough overhead imagery available with detailed textual descriptions, resulting in weak generalizability and underwhelming results for EO tasks that are out-of-distribution relative to the fine-tuning data.

\citet{zhang2024vleobench} sought to understand the extent to which the robust general knowledge of MLLMs could be transferred to the EO domain. They found that for high-level tasks, such as image captioning or scene classification, MLLMs tended to perform well. However, when tasks required more fine-grained spatial reasoning, such as object counting or localization, MLLMs did not perform well. This result aligns with work outside of the EO domain, where it has been demonstrated that MLLMs struggle with these tasks on non-overhead imagery as well \citep{kaduri2024_vision_of_vlms}. This suggests that MLLMs are capable of leveraging their vast general knowledge and applying it to EO images, but that it is the general deficiency of struggling with fine-grained visual reasoning that produces results such as those observed in \citet{zhang2024vleobench}. These challenges are only exacerbated with EO data, where tasks such as change detection \cite{ding2025survey} and object detection/segmentation \cite{hoeser2020object} require the ability to extract and contextualize small, subtle image features.

\begin{figure*}[t]
    \centering
    \includegraphics[width=\textwidth]{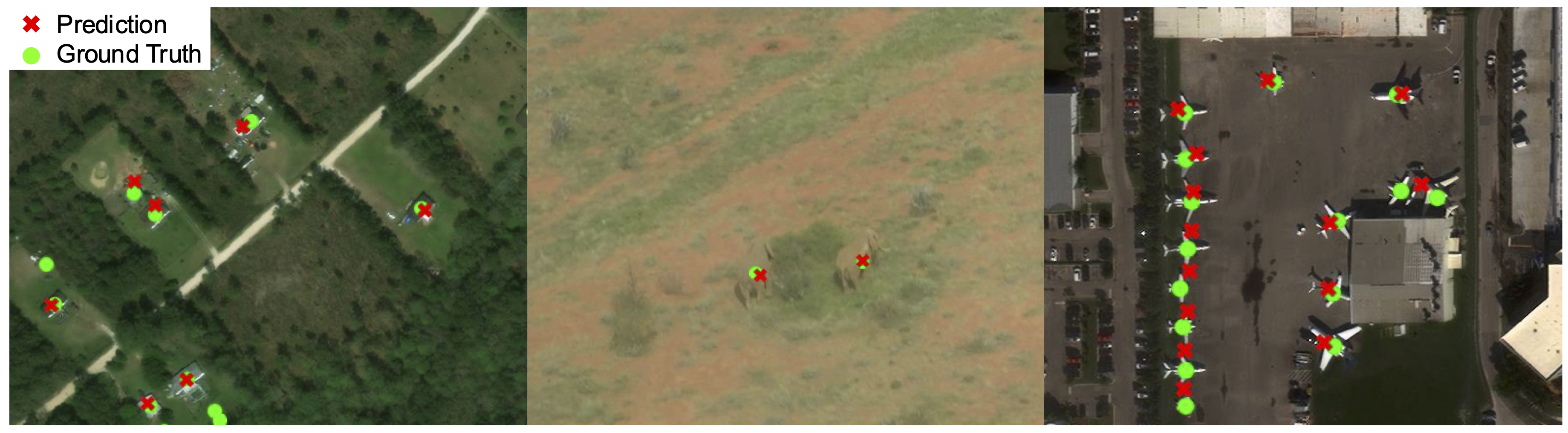}
    \caption{Sample outputs using various MLLMs (green dots=ground truth and red Xs=predictions) across three tasks: building detection (left), animal detection (middle), and plane detection (right).}
    \label{fig:example_fig}
\end{figure*}

More recent MLLMs integrate explicit localization information directly into their training pipeline \citep{deitke2024molmo,qwen2.5-VL,Qwen2VL}. These are relatively new capabilities that allow models to output object coordinates. While there are vision-language models, such as Grounding DINO \cite{liu2023grounding}, that can localize arbitrary objects given a natural language input, these models can \emph{only} localize objects. MLLMs on the other hand are capable of completing virtually any vision-language task due to their ability to autoregressively generate text tokens, making them strong candidates for an all-purpose model that can be applied to a large number of EO tasks.

We are the first work to benchmark many of these new models on the EO domain, directly assessing whether prior results regarding the poor localization capabilities of MLLMs still hold. We evaluate the Molmo family of models \citep{deitke2024molmo}, the new Qwen 2.5-VL architecture \citep{qwen2.5-VL}, and the recent Llama 3.2 model \citep{dubey2024llama}, across three different object localization tasks: plane detection, building detection, and animal detection (cropped examples of these tasks can be seen in Figure \ref{fig:example_fig}). We select Molmo and Qwen 2.5-VL because both explicitly include localization capabilities in their training pipeline, and we select Llama 3.2 as an experimental control because it was released contemporaneously with the other architectures but does not advertise localization capabilities. We begin by directly comparing the MLLMs' zero-shot localization performance to a standard object detection architecture, Faster RCNN \cite{faster_rcnn}, trained on various amounts of training data. While this is not a fully fair comparison due to the MLLMs operating in a zero-shot setting, it allows us to precisely quantify how much data is needed for a typical object detector to beat a zero-shot MLLM on EO tasks. We find that MLLMs are effective when objects are sufficiently large and distinct, in both single-class and multi-class settings. Then, we explore fine-tuning Molmo, using the same few-shot data, to determine whether fine-tuning is effective for improving domain-specific performance, finding that it increases or maintains prediction accuracy relative to the zero-shot setting. Along with these quantitative results, we provide a qualitative evaluation of each models' results via visualizations and a detailed discussion of failure scenarios, as well as other insights and lessons that we learned, including ground sample distance (GSD) optimization and prompt selection.

\textbf{To summarize, our contributions are as follows:}
\begin{enumerate}
    \item We are the first work to provide a detailed benchmark of MLLMs that are capable of outputting precise object coordinates for EO data.
    \item We evaluate MLLM performance in zero-shot and fine-tuned scenarios against standard detectors, noting that fine-tuning can enhance or sustain performance while also presenting challenges specific to EO tasks.
    \item We provide a qualitative evaluation of model behavior, highlighting specific failure scenarios and how pre-processing steps, such as GSD selection during tiling, impacts downstream performance.
    \item We detail our prompt selection strategy, illustrating how EO performance can vary significantly depending on the prompt, and providing a starting point for future work seeking to utilize these models for EO imagery.
\end{enumerate}

\section{Background}
\subsection{MLLMs with Localization Capabilities}
MLLMs with localization capabilities offer an exciting opportunity for EO, where object localization is fundamental. Training traditional object detection models, such as YOLO \cite{redmon2016you}, DETR \cite{carion2020end}, Faster R-CNN \cite{faster_rcnn}, etc., can offer strong results across many tasks, but this performance requires high quality, task-specific labeled data, which typically requires the use of domain experts with domain knowledge and can take a significant amount of time, especially if images were captured over a wide area. With strong enough out-of-the-box performance, MLLMs can completely eliminate, or significantly reduce, the need for data labeling. Or, even if annotations are needed, an MLLM can be used for data filtering, speeding up the annotation process.

In this work, we will be defining an MLLM as an autoregressive, general purpose, large language model (LLM) that includes visual inputs. We use MLLM rather than VLM (vision language model) due to the fact that VLM can be used to refer to models such as CLIP \cite{radford2021learning} or GLIP \cite{li2022grounded} that are not autoregressive. While some of these models do offer open-set object detection capabilities, such as GLIP or Grounding DINO, we believe MLLMs offer some benefits that these models do not. First, MLLMs have been exposed to much more data, including text-only data, that enhances their ability to generalize to new, unseen information. Second, MLLMs are much more flexible and are able to perform a wide array of vision-language tasks, rather than just a few localization tasks in the case of Grounding DINO or image classification/image retrieval in the case of CLIP.

MLLMs have been heavily developed over the last couple of years, with state-of-the-art models such as GPT-4o and Claude 3 including these capabilities. These models are capable of completing a wide array of vision-language tasks out-of-the-box, including image captioning, visual question answering, and more. However, most of these models \emph{are not} capable of completing standard computer vision tasks that require precise object localization, such as phrase grounding, object detection, and semantic segmentation \citep{kaduri2024_vision_of_vlms,zhang2024vleobench}. This is because the models can output relative object locations, and maybe even make an attempt at generating object coordinates, but cannot do so with the precision required for them to be broadly applied to these tasks.

Architecturally, there is nothing inherent in MLLMs that prohibit them from being used to output localization information. Indeed, some prior work has fine-tuned them to output localization coordinates in the form of text tokens \cite{zhang2023gpt4roi,kuckreja2023geochat}. However, this fine-tuning tends to sacrifice generalization at the cost of these capabilities. More recently, MLLMs have begun to appear that include localization data in their pre-training or supervised fine-tuning stages. Such models include the Qwen-VL family \cite{Qwen2VL,qwen2.5-VL} and Molmo \cite{deitke2024molmo}, which can output bounding boxes and points, respectively. Due to the fact that the localization data points are mixed with other data types during model training, these models do not suffer from the lack of generalization that is observed in MLLMs that are post-fine-tuned to include object coordinates.

\subsection{Fine-tuning MLLMs}
Directly training MLLMs can be challenging. They are prone to overfitting, require large amounts of compute, and generalization commonly degrades as task-specific performance improves. Parameter efficient fine-tuning (PEFT) methods have become the standard method of fine-tuning LLMs due to their computational efficiency and ability to be applied ad-hoc, preserving model generalizability when needed \cite{hu2022lora,liu2024dora,hanparameter}. These methods broadly work by adding additional weights to layers in the model in the form of low-rank decomposition matrices, while keeping the original weights frozen. In this work, we specifically target DoRA (Weight-Decomposed Low-Rank Adaptation) \cite{liu2024dora}, which decomposes the weights of a model into directional and magnitude components and separately trains each.

More formally, each weight matrix \( W \) is decomposed as:

\[
W = mV,
\]

where:
\( m \) is a vector of magnitudes and \( V \) is the directional component (a matrix with unit-norm columns).

During fine-tuning, both the magnitude vector \( m \) and the directional component \( V \) are adapted to the new task. The update to the directional component \( V \) is performed using two low-rank matrices \( A \) and \( B \):

\[
\Delta V = A B^\top,
\]

where: \( A \in \mathbb{R}^{d \times r} \), \( B \in \mathbb{R}^{d \times r} \), \( r \ll d \) is the rank (with \( d \) being the dimensionality of the weight matrix).

The updated directional component becomes:

\[
V^\prime = V_0 + \Delta V = V_0 + A B^\top,
\]

where \( V_0 \) is the initial directional component from the pre-trained model.

Similarly, the magnitude vector \( m \) is updated during fine-tuning using direct optimization methods. DoRA can then be fully formulated as follows:

\[
W^\prime = m(V_0 + A B^\top).
\]

DoRA significantly reduces the number of trainable parameters required during fine-tuning, enhancing computational efficiency while maintaining model performance.

\section{Experiments}
\subsection{Implementation Details}

\begin{table*}[t]
    \centering
    \begin{tabular}{l l l}
       Model & Dataset & Prompt \\\hline
        Molmo 7B O & RarePlanes & ``Where are the \{category\}?" \\
        Molmo 7B O & AAP & ``Point to \{category\}$\backslash$nPlease say 'This isn't in the image.' if it is not in the image." \\
        Molmo 7B O & xBD & ``Place a point on each \{category\} in the image." \\
        Molmo 72B & RarePlanes & ``Place a point on each \{category\} in the image." \\
        Molmo 72B & AAP & ``Look for \{category\} in the image and show me where they are." \\
        Molmo 72B & xBD & ``Place a point on each \{category\} in the image." \\
        Qwen 2.5-VL & All & ``Detect all \{category\} in the image." \\
        Llama 3.2 & All & ``Detect all \{category\} in the image. Output the coordinates in the form: [x1, x2, y1, y2]." \\\hline
    \end{tabular}
    \caption{The prompts that we used for each model to extract localization information across each task. We initially tried to match the prompts that were provided in the original papers and, in some cases, did our own additional tuning. \{category\} is the class of the target we are locating.}
    \label{tab:prompts}
\end{table*}

\paragraph{Datasets}
We evaluate the MLLMs on a variety of EO object detection tasks. The first dataset that we consider is the RarePlanes dataset \cite{shermeyer2021rareplanes}. It consists of 253 Maxar WorldView-3 satellite scenes and 14,700 aircraft annotations, with various metadata associated with each aircraft.
We make this dataset a 1 class plane detection task, assessing each MLLM's ability to detect high-level object categories. The RarePlanes scenes were tiled into 1333x800 tiles with a 200 pixel overlap. The second dataset that we consider is the Aerial Animal Population (AAP) dataset \cite{eikelboom2019improving}, which consists of a set of high resolution images taken from a helicopter with small animals belonging to one of three classes: elephant, giraffe, and zebra, assessing each model's ability to detect and classify small objects. We used the publicly available, pre-processed data from \citet{zhang2024vleobench}. The training portion of this dataset is already tiled, but we manually tiled the test portion of the dataset into 900x700 tiles with 200 overlap, approximately matching the training tiles. The last dataset that we consider is the xBD dataset. \cite{gupta2019creating} Because its images are at a relatively high GSD, this allows us to assess each models ability to localize information in large scenes. We once again use the publicly available data from \citet{zhang2024vleobench}. xBD is a satellite image dataset originally constructed for change detection. We convert it into an object detection dataset by only considering the first image in each example and asking the models to identify the buildings in the image.
To construct few-shot splits for each of our datasets, we randomly sampled $K$ images for each target category from the dataset (i.e. 3 sets of images for AAP and 1 set of images for xBD/RarePlanes), where $K$ is the number of shots. We created 10 seeds for each K-shot split, each consisting of different images and a varying number of annotations.

\paragraph{Metrics}
Molmo only outputs centerpoints \cite{deitke2024molmo}, therefore we evaluate the models in a centerpoint regime. We implement a center mAP score, where we evaluate centerpoint detections by measuring the Euclidean distance between the predicted object center $p = (p_x, p_y)$ and the corresponding ground truth center $g = (g_x, g_y)$. A predicted center is deemed a true positive if its distance from a ground truth center is below a predefined threshold $\tau$. Once true positives are determined, we compute the precision–recall curve and calculate the average precision (AP) as the area under this curve. The overall mean average precision (mAP) is obtained by averaging the AP over classes:

\[
\text{mAP} = \frac{1}{|\mathcal{C}|} \sum_{c \in \mathcal{C}} \int_{0}^{1} p_c(r) \, dr,
\]

where \( \mathcal{C} \) is the set of classes, and \( p_c(r) \) is the precision at recall \( r \) for class \( c \).

We set $\tau$ based upon the size of the target(s) in each dataset. To evaluate Qwen 2.5-VL, which outputs bounding boxes, and Faster RCNN \cite{faster_rcnn}, we convert the bounding boxes to centerpoints by considering the center of the box as the prediction. Additionally, the MLLMs do not provide direct confidence values. Therefore, when evaluating these models, we assume all confidence scores are the same, and do not include them in our metric calculation. However, for the Faster RCNN, we do include confidence scores, as this is an additional benefit of using a traditional model.

\subsection{Zero-shot Results}
\paragraph{RarePlanes}
\begin{table*}[t]
    \centering
    \begin{tabular}{l l l l}
        Model & RarePlanes mAP@30pix & AAP mAP@30pix & xBD mAP@15pix \\\hline
        Molmo 7B O & 62.62 & \textbf{30.26} & 2.97 \\
        Molmo 72B & \textbf{72.12} & 29.82 & \textbf{4.22} \\
        Qwen 2.5-VL 7B & 46.62 & 30.01 & 0.49 \\
        Qwen 2.5-VL 72B & 50.03 & 12.09 & 0.50 \\
        Llama 3.2 11B & 0.00 & 0.00 & 0.00 \\
        Llama 3.2 90B & 0.00 & 0.00 & 0.00 \\\hline
    \end{tabular}
    \caption{Object detection results for various MLLMs across three different datasets.}
    \label{tab:obj_det}
\end{table*}
\begin{figure*}
    \centering
    \includegraphics[width=\textwidth]{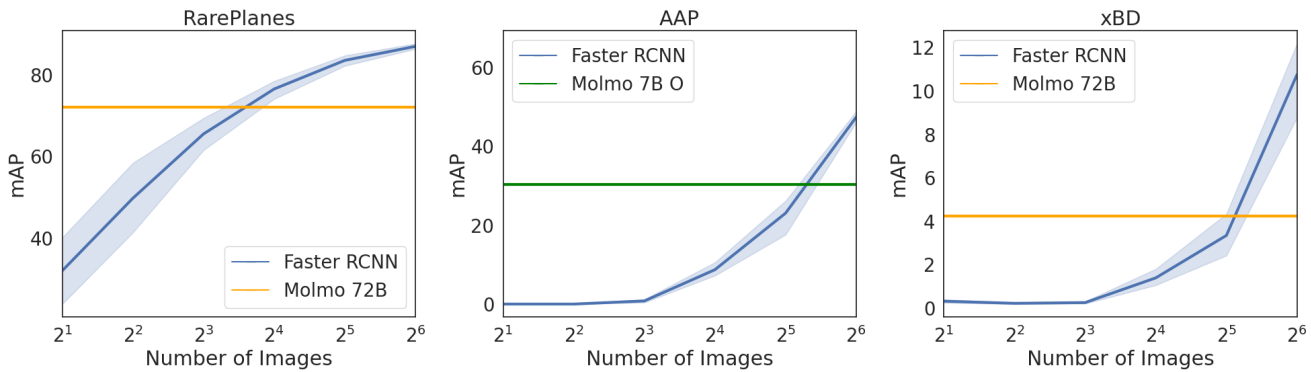}
    \caption{Few-shot Faster RCNN \citep{faster_rcnn} performance with varying amounts of training images (blue lines) vs. top-performing MLLM's performance (alt-color lines) for each task.}
    \label{fig:faster}
\end{figure*}

The second column of Table \ref{tab:obj_det} contains the results for each MLLM on the RarePlanes plane detection task. This is the easiest task out of the three due to the size and distinct shape of the targets. We find that Molmo 72B performs the best out of all of the MLLMs. Note that both Llama models obtain a mAP of 0 due to the fact that all of its outputs are invalid. Even after exploring a variety of prompting strategies, it was difficult to get the model to output anything reasonable (see examples in Supplementary). This suggests that Llama was never exposed to this type of data during training and, despite getting strong performance on MLLM benchmarks, cannot easily generalize to localization tasks. Qwen 2.5-VL is the newest of the models, however, both of the Molmo models outperformed it.

Next, we compare the most performant MLLM, Molmo 72B, to a Faster RCNN, trained on randomly sampled subsets of data. Figure \ref{fig:faster} (left) illustrates the performance observed as more data is added to the Faster RCNN's training dataset, while the horizontal line represents the performance of Molmo 72B. We can see that the Faster RCNN performance intersects with Molmo between 8 and 16 examples, indicating that this is the number of images needed to beat the zero-shot performance of Molmo on this task. This suggests that when a task is relatively simplistic and even just a few labeled examples are available, it may not be worth leveraging an MLLM. Instead, it is best to leverage a specialized object detection architecture like Faster RCNN or YOLO. However, the performance that Molmo achieves on this task is impressive for a zero-shot model and, in data scarce settings, it is still worth considering. Various visualizations are available in the supplementary, illustrating successes as well as failures, including producing false positives on shadows, missing a plane when it is close to another, and missing planes that are partially obscured.

\paragraph{Aerial Animal Population}
The third column of Table \ref{tab:obj_det} contains the results for each MLLM on the AAP animal detection task. This task is more challenging than RarePlanes due to the objects of interest being smaller and it being a multiclass problem. Due to this difficulty, each of the models perform worse relative to RarePlanes. Llama once again is incapable of localizing any of the objects in the scene. Molmo 7B O outperforms all other models on this task, including Molmo 72B by a small margin. Even more surprisingly, Qwen 2.5-VL 7B substantially outperforms its 72B counterpart. This suggests that rather than the size of the model predicting performance, there is likely a complex interaction between the original training data and the number of parameters in the model that determines its performance for a given task, and that all models within a model family, not just the largest, should be explored for each new task. For most models, `elephant' has the highest AP, likely due to its size and the fact that it is less likely to be confused with the other targets.

Figure \ref{fig:faster} (middle) contains results for Faster RCNN on this task. Like the MLLMs, Faster RCNN struggles with the task compared to RarePlanes, despite being explicitly designed for object detection. Notably, its performance does not intersect Molmo 7B until 32 examples are included during training. While acquiring 32 labeled images may be trivial for some tasks, this task is a great illustration of a scenario in which this might be challenging. Namely, the targets are relatively small, making them easy to miss for a human annotator, and, when tiled, there are a large number of images without targets.
While some scenarios require high accuracy and labeling, others do not, making MLLMs attractive. They can be useful for site monitoring when labeled data is unavailable, with further refinement using a traditional classifier for detailed categorization.
Figures illustrating the performance of Qwen on this dataset, along with detailed descriptions of various success and failure cases, are available in the supplementary. In many images, Qwen performs quite well, correctly identifying most of the animals. However, there are catastrophic failure scenarios. For instance, in some cases, Qwen only places a single box on a group of animals, hurting average precision.

\begin{table}[t]
    \centering
    \begin{tabular}{l l l l}
        Molmo 7B Model & RP mAP@30pix & AAP mAP@30pix \\\hline
        Zero-shot &  59.83 & 29.20 \\
        RP Fine-tune & 62.03 & 28.69 \\
    \end{tabular}
    \caption{Object detection results for Molmo 7B D\protect\footnotemark, comparing zero-shot performance and fine-tuned performance. The fine-tuned model was trained on RarePlanes (RP) and evaluated on both RP and AAP, testing in-distribution and out-of-distribution performance respectively.}
    \label{tab:finetune}
\end{table}

\paragraph{xBD}
The fourth column of Table \ref{tab:obj_det} contains the results of each MLLM on the xBD building detection task. None of the models perform particularly well at the task. The Molmo family of models is the only one that outputs some valid predictions. Examples are included in the supplementary illustrating the challenging nature of this task. When buildings are sufficiently large, the model is able to accurately detect them. However, there are many small buildings in the image that are difficult to discern without zooming into the image, making this a challenge for all models, and even humans without zooming in and carefully scanning each image. The Faster RCNN performance is visible in Figure \ref{fig:faster} (right). Even a trained detector does not perform well when up to 128 examples are included in the training set; this is another testament to the task's difficulty.

\subsection{Fine-tuning Results}

Based upon the prior experiments, we determined that a Faster RCNN begins to outperform Molmo 7B with 16-64 examples for RarePlanes, AAP, and xBD. However, in each of these settings, Molmo was operating in a zero-shot fashion. Molmo is not an overhead specific model, and while some EO data was likely included during pre-training, there is still a domain gap between the majority of images that Molmo was trained on and each of these datasets. Therefore, we explored the degree to which Molmo's performance can be improved by fine-tuning in a few-shot setting. Namely, we focus on the RarePlanes dataset and train Molmo 7B with 16 examples, the same amount where the Faster RCNN surpasses zero-shot performance. 

\footnotetext{At the time, only Molmo 7B D was available for fine-tuning}

The results of this experiment can be found in Table \ref{tab:finetune}. We found that fine-tuning, even at such a low-shot regime, was effective for improving performance on this task, albeit if only slightly at $+2.20$ mAP. We suspect more examples are needed to increase PEFT fine-tuning efficacy on this task, along with a deeper exploration into the trade-off between performance and generalization as impacted by DoRA's rank and alpha, and the number of total training examples. Improvements are also likely to be found in the targeted application of adapter modules, specifically by targeting layers in the model that are correlated with larger improvements in performance. While fine-tuning on a specific dataset obviously improves performance on that task, generalization often suffers as a result of this fine-tuning. To verify this in our setting, we evaluate the same RP fine-tuned model on the AAP dataset. Note that while this is still overhead object detection, it is much different than RP; it consists of imagery from a helicopter, rather than satellite imagery, and the targets are substantially different. The results on this out-of-distribution evaluation task are also available in Table \ref{tab:finetune}. Unfortunately, we do observe a performance drop on this dataset of $-0.51$ mAP. Overall, this suggests that our standard fine-tuning setup slightly harmed the performance of Molmo in this few-shot setting. However, such small degradations across 10 experiments could still be within the margin of error of the 0-shot results. 

These results highlight the difficulties in fine-tuning MLLMs. We hope that future work will continue to investigate effective methods for fine-tuning these models for EO tasks. Our findings suggest that large foundation models may be particularly useful for rare object detection with very few labeled examples and for broad-area search when deploying models globally to locate targets anywhere. Additionally, MLLMs appear well-suited for adaptation into all-purpose EO detectors given the incremental degradation observed on out-of-domain datasets.\footnote{Note that as more localization data is included in training these models, one may achieve strong enough performance to be an ``all-purpose" EO detector.} Due to the nature of PEFT-based methods, including DoRA, it is easy to swap out and even mix adapter modules for specific tasks, allowing flexibility in applying these models to different scenarios that may or may not require fine-tuning.
\cite{hu2022lora, liu2024dora, cai2024survey,muennighoff2024olmoe}.

\subsection{Additional Analyses}
\paragraph{Prompt Tuning}
\begin{figure}
    \centering
    \includegraphics[width=\columnwidth]{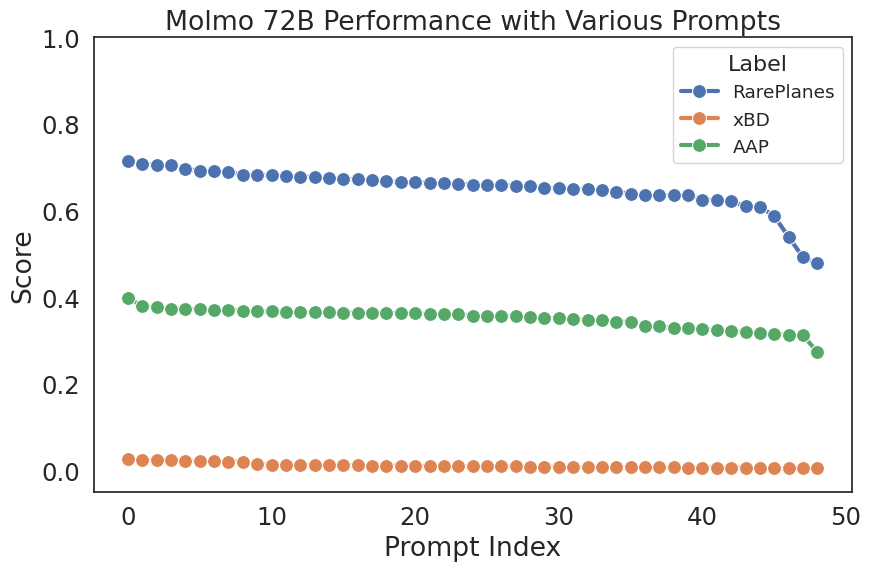}
    \caption{RarePlanes mAP scores across 50 prompts using Molmo 72B and evaluating on 50 random examples.}
    \label{fig:prompt_lg}
\end{figure}

\begin{figure}
    \centering
    \includegraphics[width=\columnwidth]{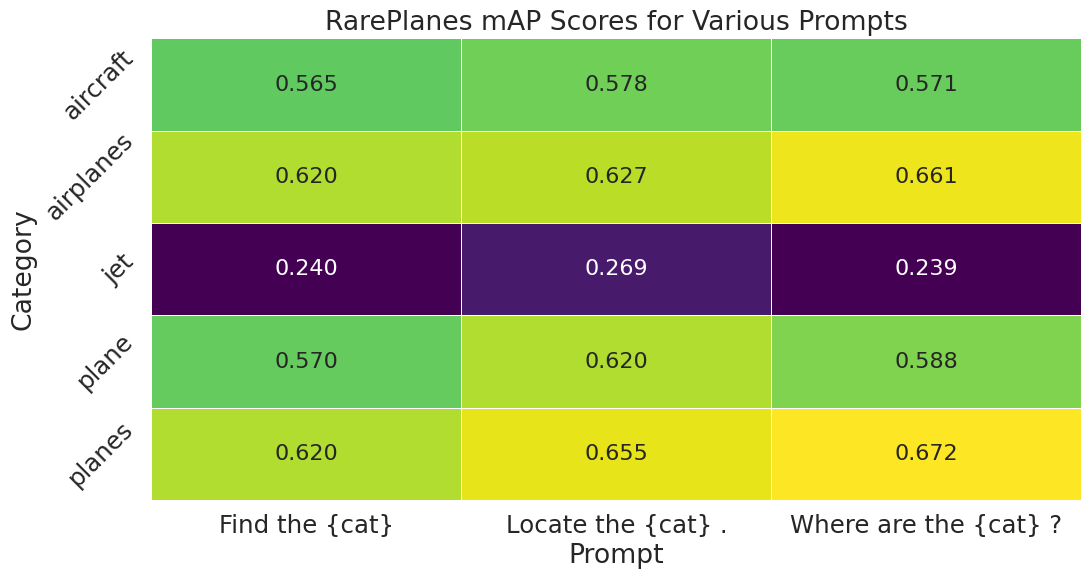}
    \caption{RarePlanes mAP scores for various prompts using Molmo 7B O and evaluating on 200 random examples. \{cat\} is the location where each of the categories on the y-axis are inserted.}
    \label{fig:prompt_cm}
\end{figure}

MLLMs accept natural language queries, resulting in a large number of possible prompts that convey the same information. However, prompting can substantially impact downstream model performance, with some prompts outperforming others, seemingly inexplicably. For each of the models that we implemented, we began with prompts that were provided in the original papers (if prompts/examples were made available), and conducted a search. We conducted the most thorough search using Molmo. Molmo provides the full list of prompts that were used for training, providing a set of 49 prompts that we could sample from. Some example prompts can be seen in Figure \ref{fig:prompt_cm} on the x-axis. For each of our datasets, we evaluated each of the 49 prompts on a randomly sampled selection of 50 images. We then selected the highest performing prompt and used this for evaluation on the test dataset. The performance distribution across each of these 49 prompts can be seen for Molmo 72B in Figure \ref{fig:prompt_lg}. The range of mAP across the prompts is 23.69 for RarePlanes, with the best prompt being `I am looking for airplane, where can they be found in the image?' and the worst prompt being `Find a ``airplane".' Unfortunately, there is nothing about either prompt that suggests its respective performance on the task. One might think that the poor performance of the latter prompt could result from the fact that it is referring to only a single airplane, while most images contain multiple airplanes. However, the third worst prompt is `Point to all occurrences of ``airplane"', scoring just 6 mAP higher. Surprisingly, we found that a custom prompt that we had initially explored, seen in Table \ref{tab:prompts}, outperformed the prompts that were used during training. This suggests that a large search space is warranted for EO imagery and that Molmo generalizes beyond the prompts that it was trained on.

We then conducted a more thorough prompt search for Molmo 7B O on the RarePlanes dataset. We selected the 3 top scoring prompts from our prior search and conducted an additional search over potential categories that can be used to refer to the targets in RarePlanes. We started with the word ``airplane" and used word2vec \cite{mikolov2013efficient} to generate 5 synonyms by selecting words with the closest cosine similarity. We evaluated each of these 5 categories using the 3 top scoring prompts. The results of this experiment can be found in Figure \ref{fig:prompt_cm}. The target categories lend themselves to a more intuitive interpretation than the prompts. ``Airplanes" and ``planes" perform the best; we hypothesize that this is due to the fact that they are plural. Most of the images in the RarePlanes dataset that contain planes, contain multiple planes. This likely biases the models towards predicting multiple planes, boosting performance in this case. ``Aircraft" is the second worst category because this is a hypernym for ``airplane". The RarePlanes dataset includes targets such as helicopters, which are unlabeled distractors, and we noticed that ``aircraft" tends to produce false positives on them. Lastly, ``jet" performs the worst out of all the categories that we explored. In this case, ``jet" is a hyponym and produces false negatives on some of the targets.

\paragraph{Optimal GSD Exploration}
\begin{figure*}
    \centering
    \includegraphics[width=\textwidth]{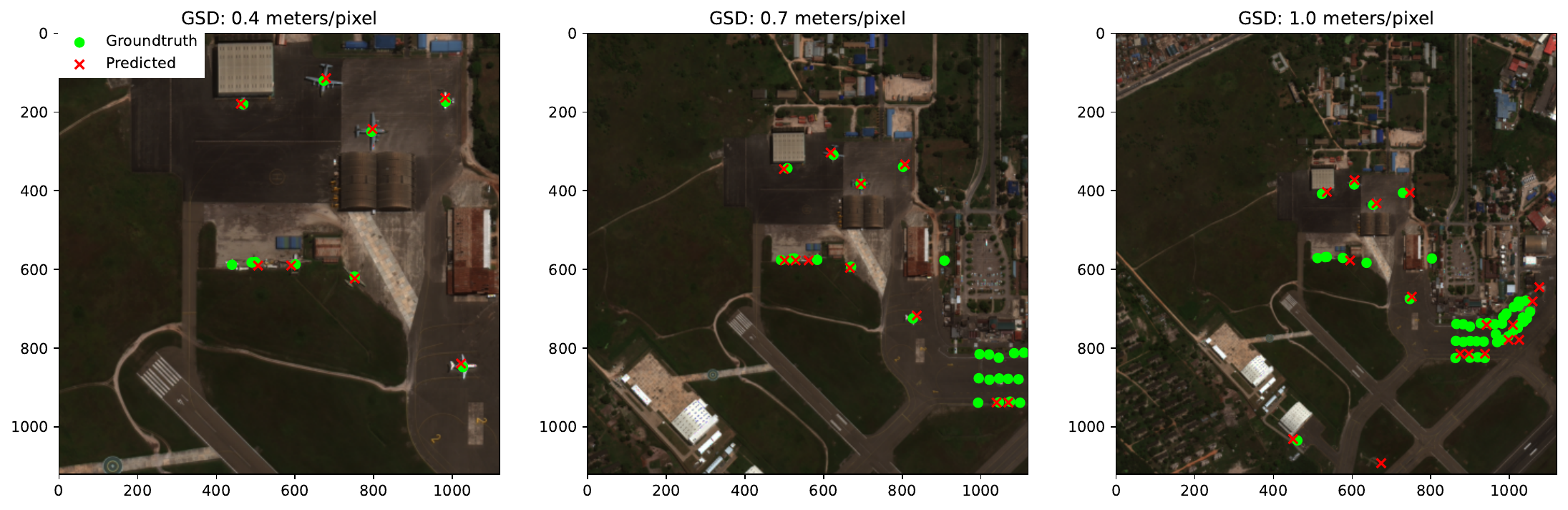}
    \caption{Tiles from the RarePlanes dataset created at various GSDs. Each image is scaled and cropped to 1120 pixels by 1120 pixels, but the spatial extent increases and objects become less resolved at higher GSDs.}
    \label{fig:gsd_ex}
\end{figure*}

\begin{figure}
    \centering
    \includegraphics[width=\columnwidth]{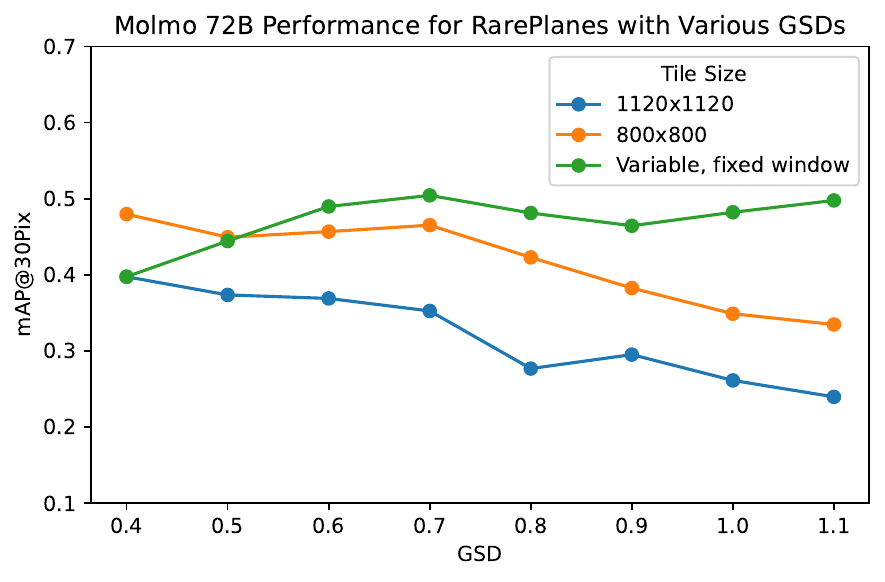}
    \caption{RarePlanes mAP@30 pixels across various rescaled GSDs for 1120x1120 pixel tiles, 800x800 pixel tiles, and a variable tile size that decreases with increasing GSD in order to maintain a fixed spatial extent.}
    \label{fig:gsd_results}
\end{figure}

GSD significantly impacts model performance in EO tasks. Models trained on one GSD range often struggle to generalize to others, especially with minimal overlap. MLLMs likely encounter a narrow GSD range during training, limiting their adaptability. To assess this, we tested performance variations using the RarePlanes dataset, which allows exploration across a wide GSD spectrum. 
We re-sampled images from their native 0.13–0.39 meters per pixel resolution to GSDs between 0.4 and 1.1 meters per pixel at 0.1-meter intervals using bi-linear interpolation. After re-scaling, we tiled images into 1120×1120, 800×800, and variable-sized tiles (adjusted to maintain a fixed spatial content window as GSD increased). Figure \ref{fig:gsd_ex} shows sample tiles, highlighting differences in object size and scene coverage.

The results of our GSD experiment using Molmo 72B can be seen in Figure \ref{fig:gsd_results}. For a fixed tile size of 1120x1120 pixels, mAP generally decreases for higher rescaled GSDs. There are a number of reasons that this might be the case. For one, the model might perform better when it has a certain amount of context surrounding the objects that it is detecting, providing important contextual information that enhances localization. Alternatively, perhaps most of the overhead training images that the model was exposed to were within a typical GSD range around 0.3 meters/pixel and, while the model does generalize, it still performs better at this specific GSD. We explored performance related to surrounding context by using a smaller tile size. For a fixed tile size of 800x800 pixels, mAP similarly decreases for higher GSDs but is nearly 10\% higher than the model's performance using 1120x1120 pixels tiles. When we varied the tile size but fixed the spatial content, mAP remains somewhat constant and higher than either the 1120x1120 or 800x800 pixel tiles after 0.6 meters per pixel GSD. These results suggest that object resolution and the amount of spatial content surrounding the object are important factors to consider when applying one of these models to a new task, rather than just using the smallest GSD available or large tiles to reduce inference time. 
Notably, our mAP scores here are lower than in Table \ref{tab:obj_det}, likely due to differences in tiling strategies, prompts, and rescaled GSDs.

\paragraph{Failure Scenarios and Limitations}
While each of the MLLMs perform differently across each of the tasks, they share some common failure scenarios (outside of Llama 3.2 which fails to generate valid coordinates for almost every example). Across all tasks, models tend to produce more false negatives than false positives. Objects are more likely to be missed if they are very small, obscured, or close to other objects. One explanation for some of these failure cases is the fact that none of the models place points with extreme precision, leading the model to think that it placed a point on an object when in fact it was meant for the adjacent one. Another reason that might explain this propensity towards false negatives is that unlike traditional detection architectures, there is no notion of an object score; the model itself is deciding which regions in the scene most strongly correspond to the desired target category. If a score was available, or we were able to lower the model's threshold for determining whether something was an object, some false negatives would be avoided. In some cases, false negatives result from too many objects being contained in the image. We notice that all of the models can only scale to a certain number of objects, likely due to the distribution of object counts in the training data. Beyond this number, the model stops generating predictions, even with sufficient output tokens. Generally, false positives tend to align with reasonable distractors. For instance, a large boulder or tree that looks like an elephant may be mistaken for one, or one target is missed within a tightly packed group of targets. However, there is an exception where the model will generate a long string of false positives (see Figure 1 in the supplementary), resulting in severe failure. We hope that future work may investigate these hallucinations further, better understanding and mitigating them.

\section{Conclusion}
In this work, we analyzed a suite of recent MLLMs across a variety of EO object detection tasks. We found that in certain scenarios, where objects are relatively distinct and are of sufficient size, that these models are capable of obtaining strong results. We then compared their performance to a standard object detector, precisely quantifying the amount of training data needed to match their performance. We provided a detailed discussion of various tips and tricks that we learned by applying these models to EO imagery, including prompt selection, GSD optimization, and failure mode avoidance. Explicit object localization capabilities are relatively new to MLLMs, offering exciting possibilities as they are further developed. Large foundation models may be particularly valuable for rare object detection with minimal labeled data and broad-area search when deploying models globally. Some challenges that remain include improved fine-tuning strategies for domain-adaptation, the development and evaluation of methodologies for improving MLLM efficiency, and the adaption of MLLMs to non-RGB imagery.
As MLLMs are applied to new tasks, such as computer control \citep{OpenAI2023OperatorSystemCard,anthropic_computer_use}, fine-grained spatial reasoning capabilities will become more important to the companies that train these models. We believe that there will be an increased interest in localization moving forward and that the EO community will be able to benefit as new models are released. We hope that this work will get others interested in these models and encourage further work focused on leveraging and understanding them.

\section*{Acknowledgments}

The research described herein was funded by the Generative AI for Science, Energy, and Security Science \& Technology Investment under the Laboratory Directed Research and Development Program at Pacific Northwest National Laboratory (PNNL), a multiprogram national laboratory operated by Battelle for the U.S. Department of Energy. This work was also supported by the Center for AI.

{
    \small
    \bibliographystyle{ieeenat_fullname}
    \bibliography{main}
}

\clearpage
\maketitlesupplementary

\section{Implementation Details}
\paragraph{MLLMs}
We used the HuggingFace implementations for each of the MLLMs \citep{wolf-etal-2020-transformers} and opted to use the instruct variants of each model. For Molmo, we used the standard release of Molmo 7B O and a 4-bit quantized version of Molmo 72B. For Qwen 2.5-VL 7B, we used the public code release and we used the 4-bit quantized version of Qwen 2.5-VL 72B. For Llama 3.2 11B, we used the original release by Meta and we used a 4-bit quantized version of Llama 3.2 90B. All MLLMs were run on a single Nvidia H100 GPU. We set the max new tokens to 800 tokens. We used the default settings for Llama 3.2 and Qwen 2.5-VL and for Molmo we specified a greedy sampling strategy for token generation. The prompts that we used for each model in our zero-shot and fine-tuning experiments are available in Table 1 in the main paper.

\paragraph{Few-shot Details}
To construct few-shot splits for each of our datasets, we randomly sampled $K$ images for each target category from the dataset (i.e. 3 sets of images for AAP and 1 set of images for xBD/RarePlanes), where $K$ is the number of shots that we are considering. We created 10 seeds for each K-shot split, each consisting of different images and a varying number of annotations. For RarePlanes, we trained our Faster RCNN using Detectron2 \citep{wu2019detectron2} for RarePlanes. The Faster RCNN has a ResNet-50 \citep{he2016deep} backbone and we used the ResNet 50 weights provided by Detectron2 to initialize our model. We used the standard configuration provided by Detectron2, modifying the images per batch to 2, the base learning rate to 0.0025, and the ROI head batch size to 64. We trained each model on an Nvidia A100 GPU for 1000 iterations, as we find this sufficient for convergence. For AAP and xBD, we used MMDetection \cite{mmdetection} to train the Faster RCNNs. We used the standard COCO configuration for MMDetection and the available ResNet 50 pre-trained weights to initialize the model. \par 

Few-shot experiments were conducted using DoRA \cite{liu2024dora} with rank$=8$ and $\alpha = 16$. Molmo 7B-D models were trained for 12 epochs at a learning rate of $1e-5$ on each 16-shot task. DoRA weights were merged only for inference, with all inferencing being conducted on the full datasets. Only the latest model checkpoint was saved and used for inferencing. We allowed adaptations to both Molmo's language and vision components. We suspected that false positives in the 0-shot model could be remedied by tailoring model output to a few concise centerpoint annotations. Given Molmo's apparent lack of exposure to overhead imagery, the vision component was adapted on in order to enrich its overhead feature space and allow for slight domain shift. For response generation, top-k, top-p, temperature, and sampling were not used in order to limit the verbosity of the responses and in turn, catastrophic hallucinations. \par 
To produce trainable annotations, we convert the original coco-formatted annotations of the RarePlanes \cite{shermeyer2021rareplanes} and Aerial Animal Population \cite{eikelboom2019improving} data sets to VQA-style annotations \cite{liu2023visualinstructiontuning} to be ingested by the model. To construct annotations aligning with Molmo's \cite{deitke2024molmo} training strategy, we randomly sample from the template VQA point prompts it provides. That is, each annotation's prompt comprises our label injected into template point prompts seen by Molmo during its original training. 

\paragraph{GSD Exploration Details}
To generate the tiles for our various GSDs across the RarePlanes dataset, we get the original row and column GSDs from the original image metadata and re-scale each image using bi-linear interpolation along both dimensions to match our target GSDs from 0.4 meters per pixel to 1.1 meters per pixel. We tile these re-scaled images with 200 pixels of overlap and only keep tiles where at least one annotation bounding box is at least 50\% contained in a tile. If the re-scaled image is smaller than the specified tile size, that image is skipped in our resulting set of tiles for that GSD. Finally, for the set of tiles with variable tile size for fixed spatial content, we take the initial 1120 pixels by 1120 pixels per tile at 0.4 meters per pixel and re-scale each image to each of the larger GSDs. For each tile-size and GSD, we used Molmo 72B with max new tokens set to 600 tokens, providing each tile and text prompt ``Count the number of aircraft in the image." We then use the centerpoints from the centers of the re-scaled bounding boxes to compare to the model's predicted output centerpoints, collapsing all labeled aircraft objects in the RarePlanes annotations into one ``aircraft" category.

\section{Llama Model Outputs}
As mentioned in the main paper, Llama 3.2 is unable to output valid coordinates, despite the prompt explicitly asking it to do so. Here are some sample outputs that it gave us:
\begin{enumerate}
    \item The image shows a runway and a parking lot, with four airplanes visible. \\ The coordinates of the airplanes are: [0.000000, 0.000000, 0.000000, 0.000000] [0.000000, 0.000000, 0.000000, 0.000000] [0.000000, 0.000000, 0.000000, 0.000000] [0.000000, 0.000000, 0.000000, 0.000000]
    \item There are 3 airplanes in the image. The coordinates of the airplanes are: [1] [1017, 1221, 1043, 1264] [2] [1022, 1213, 1051, 1256] [3] [1048, 1226, 1057, 1271]
    \item $[0, 0, 0, 0.5] [0.85, 1, 0.5, 1]$
\end{enumerate}

It does give valid outputs that are able to be parsed. However, the coordinates frequently make no sense in reference to the underlying image. Most of the time, it scales the coordinates from the range 0 to 1. However, even when it does output coordinates in this format, it frequently does not correspond to valid locations in the image. For instance, the third output looks like it might be valid, however, these points do not actually correspond to the planes in the image. The most common output that we observed was similar to the first output, where the model might get the correct number of planes (or might not) and will simply output a sequence of zeros for the coordinates, resulting in a complete failure scenario. As we made adjustments to the prompt to try to circumvent these issues, we found that Llama was difficult to adapt and had trouble following more detailed system prompts. Note that these issues to do preclude Llama from being used for other tasks that do not require precise localization output, such as object counting, scene description, etc. However, our experiments indicate that without fine-tuning, Llama is not a reliable localization model.

\section{Additional Figures}
The remainder of the figures in the supplementary illustrate specific success and failure scenarios for various models across each of the datasets.

\begin{figure*}[b]
    \centering
    \includegraphics[width=\textwidth]{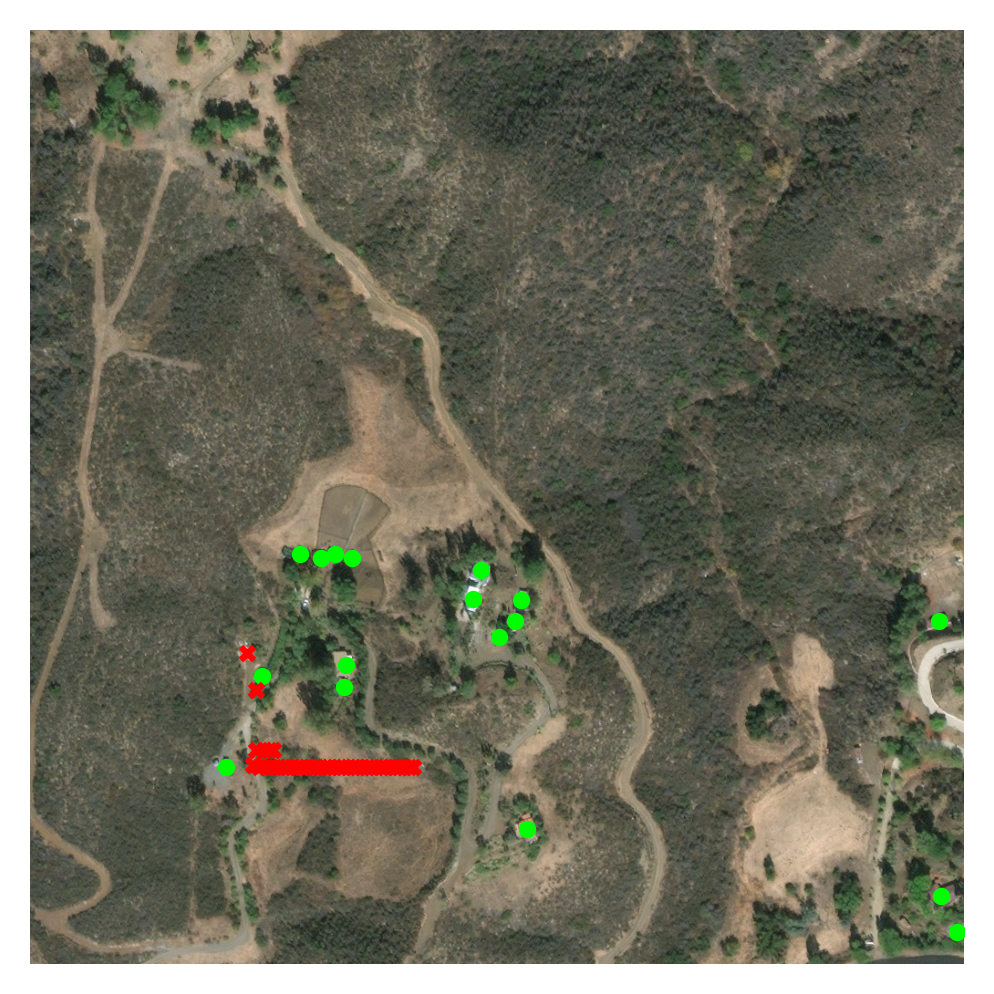}
    \caption{Image from xBD dataset with Molmo 72B labels (ground truth represented by green dots and predictions represented by red Xs). This illustrates the common failure scenario that is discussed in the main text, where models will sometimes generate a sequence of many detections in a line. We are uncertain what results in this behavior but we notice it more with small models.}
    \label{fig:general_failure}
\end{figure*}

\begin{figure*}[b]
    \centering
    \includegraphics[width=\textwidth]{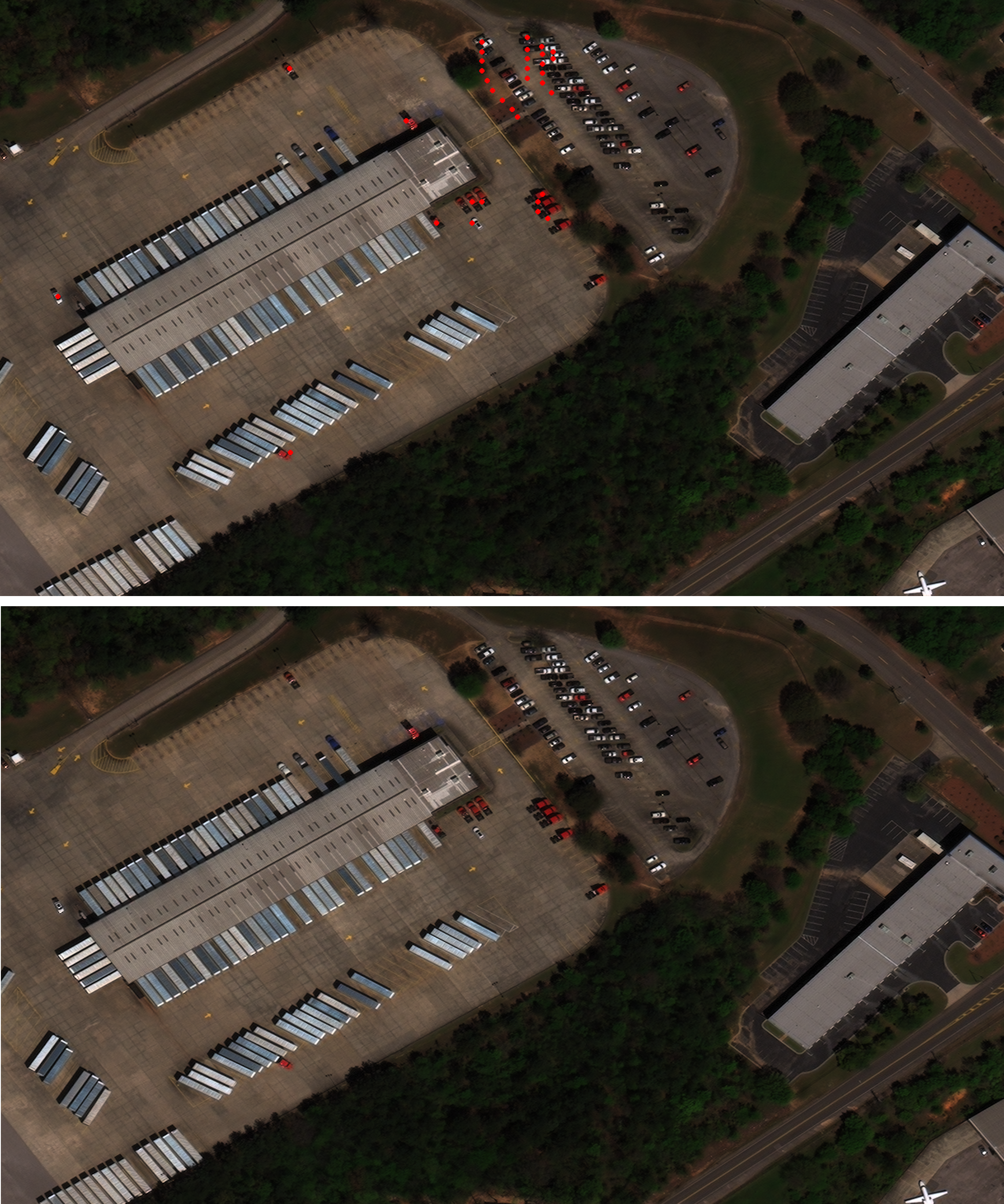}
    \caption{Above: hallucinations resulting from using the prompt ``Place a point on each \{category\} in the image", with a top-p$=0.9$, top-k$=50$, and temperature$=0.6$. Below: reduced hallucinations resulting from using the prompt ``Where are the \{category\}?", and without using top-p, top-k, and temperature. Red dots indicate predictions.}
    \label{fig:reduce_hallucinations}
\end{figure*}

\begin{figure*}[h]
    \centering
    \includegraphics[width=\textwidth]{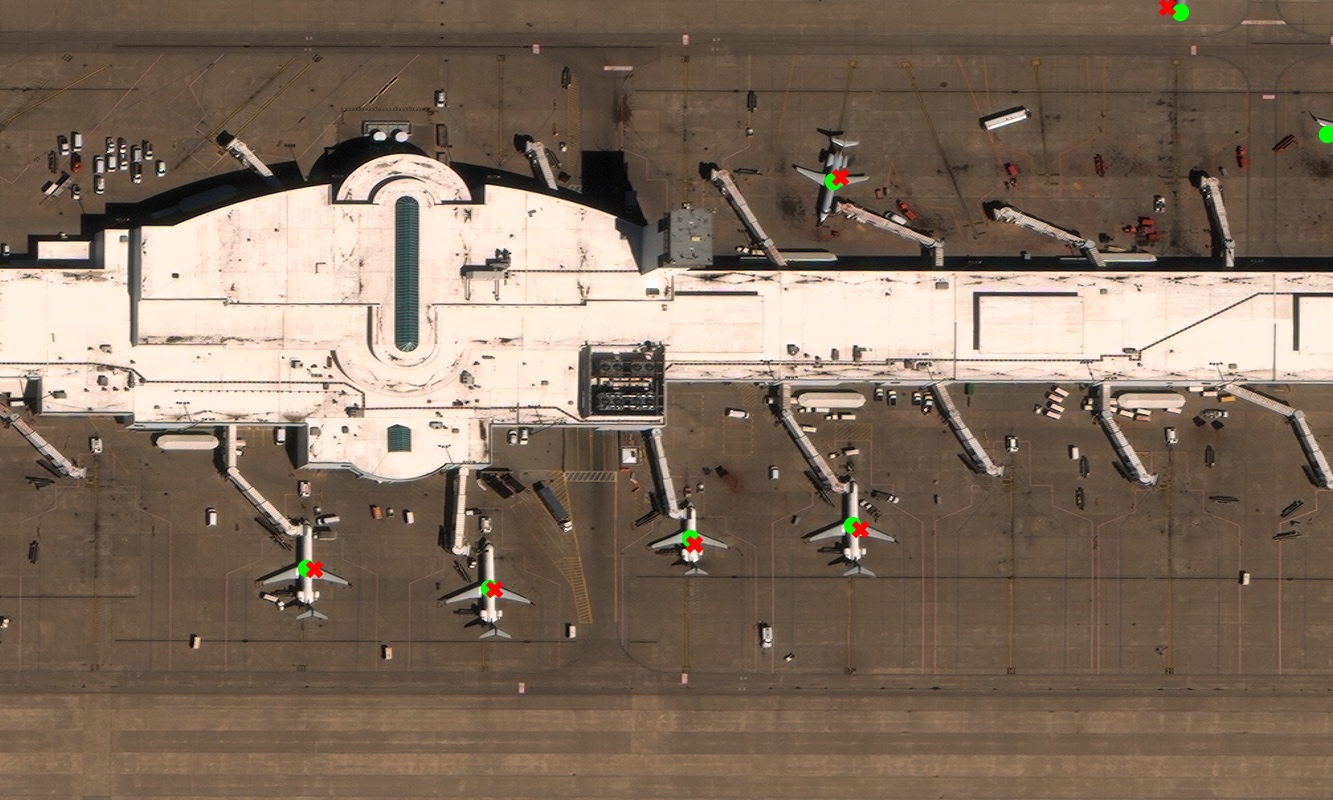}
    \caption{Illustration of an example from RarePlanes using Molmo 72B (ground truth represented by green dots and predictions represented by red Xs). Here, the model successfully detects most aircraft in the scene, despite the terminal providing many distractors. It even detects one of the two aircraft that appear at the edge of the image, suggesting that it is able to detect parts of planes.}
    \label{fig:rp_success}
\end{figure*}

\begin{figure*}[h]
    \centering
    \includegraphics[width=\textwidth]{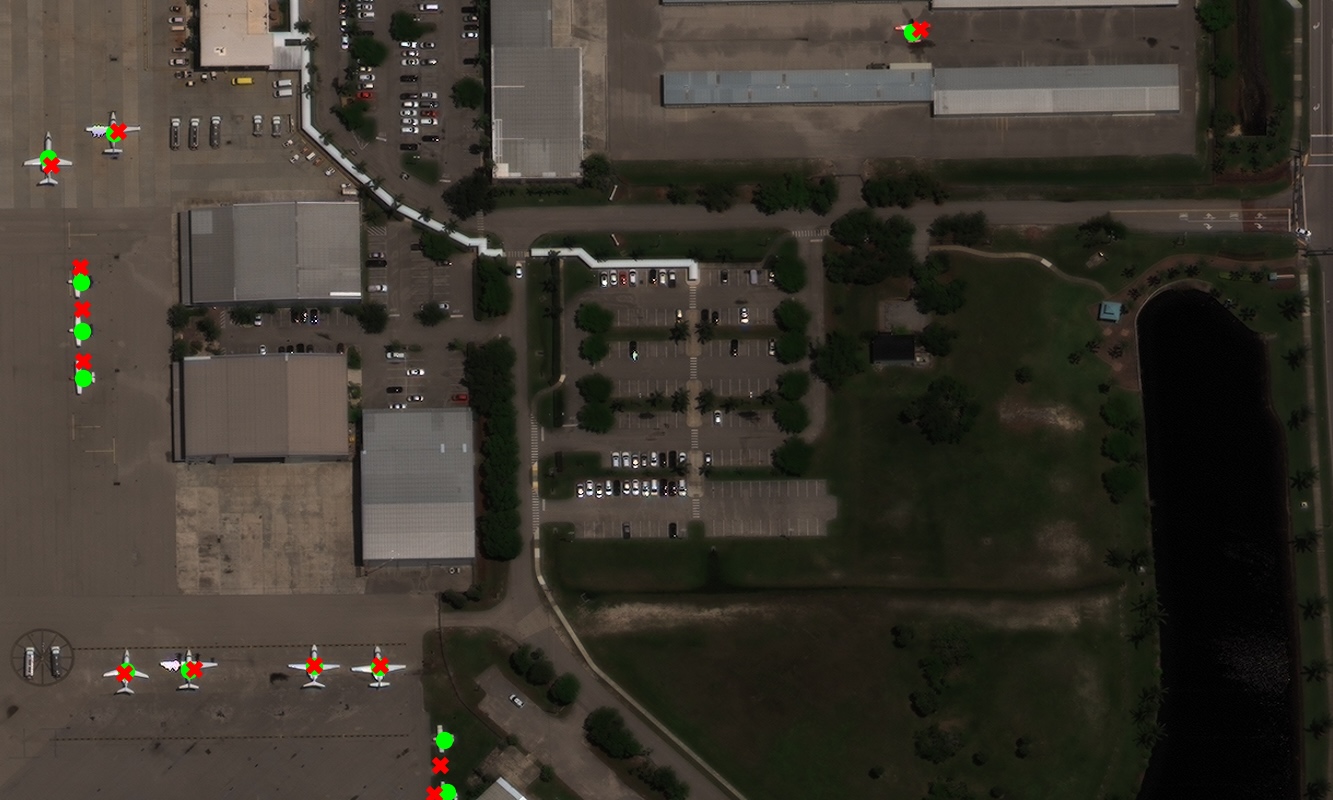}
    \caption{Illustration of an example from RarePlanes using Molmo 72B (ground truth represented by green dots and predictions represented by red Xs). Here, the model successfully detects all aircraft in the image, despite variations in size and orientation.}
    \label{fig:rp_success_2}
\end{figure*}

\begin{figure*}[h]
    \centering
    \includegraphics[width=\textwidth]{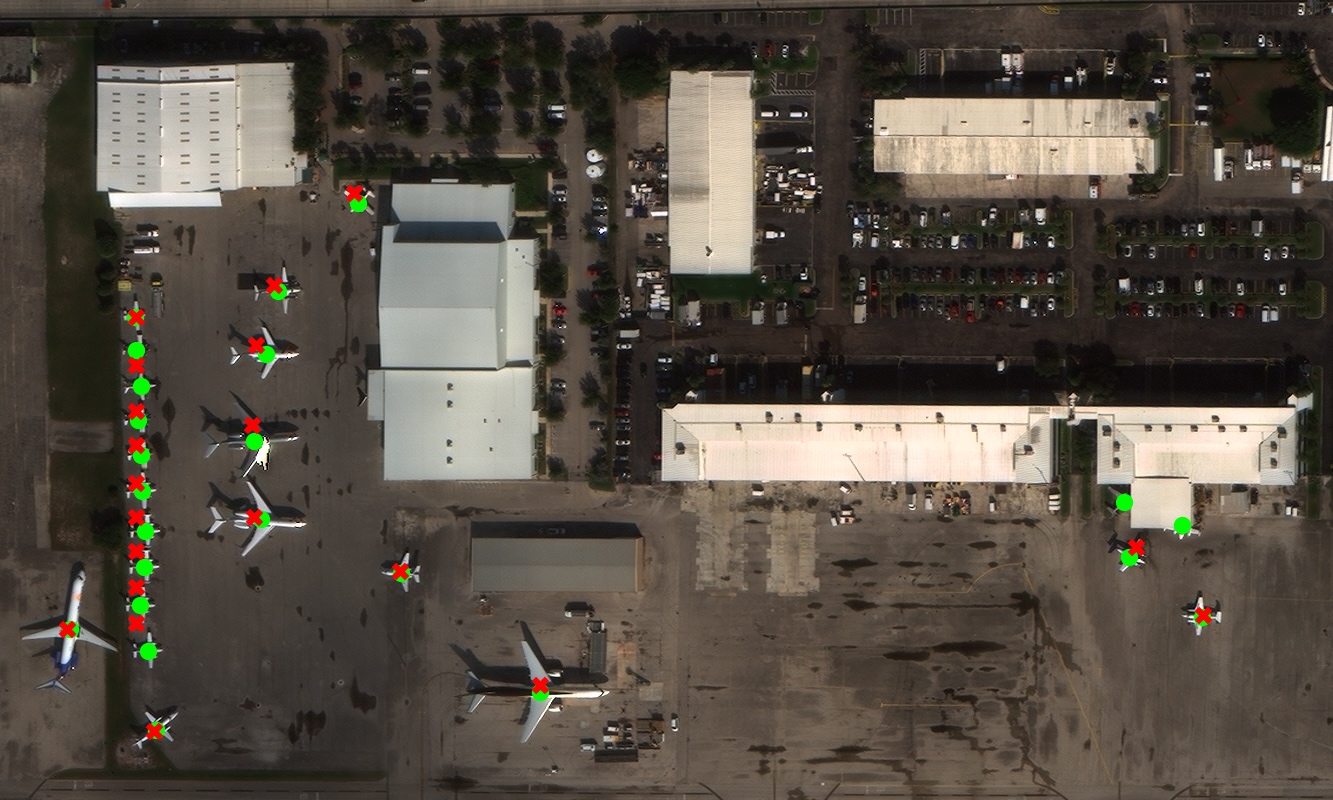}
    \caption{Illustration of an example from RarePlanes using Molmo 72B (ground truth represented by green dots and predictions represented by red Xs). The model successfully predicts most planes in the image, despite the large number of targets and potential distractors. It misses a plane that is in close quarters to other planes and it misses two planes that are partially obscured.}
    \label{fig:rp_failure}
\end{figure*}

\begin{figure*}[h]
    \centering
    \includegraphics[width=\textwidth]{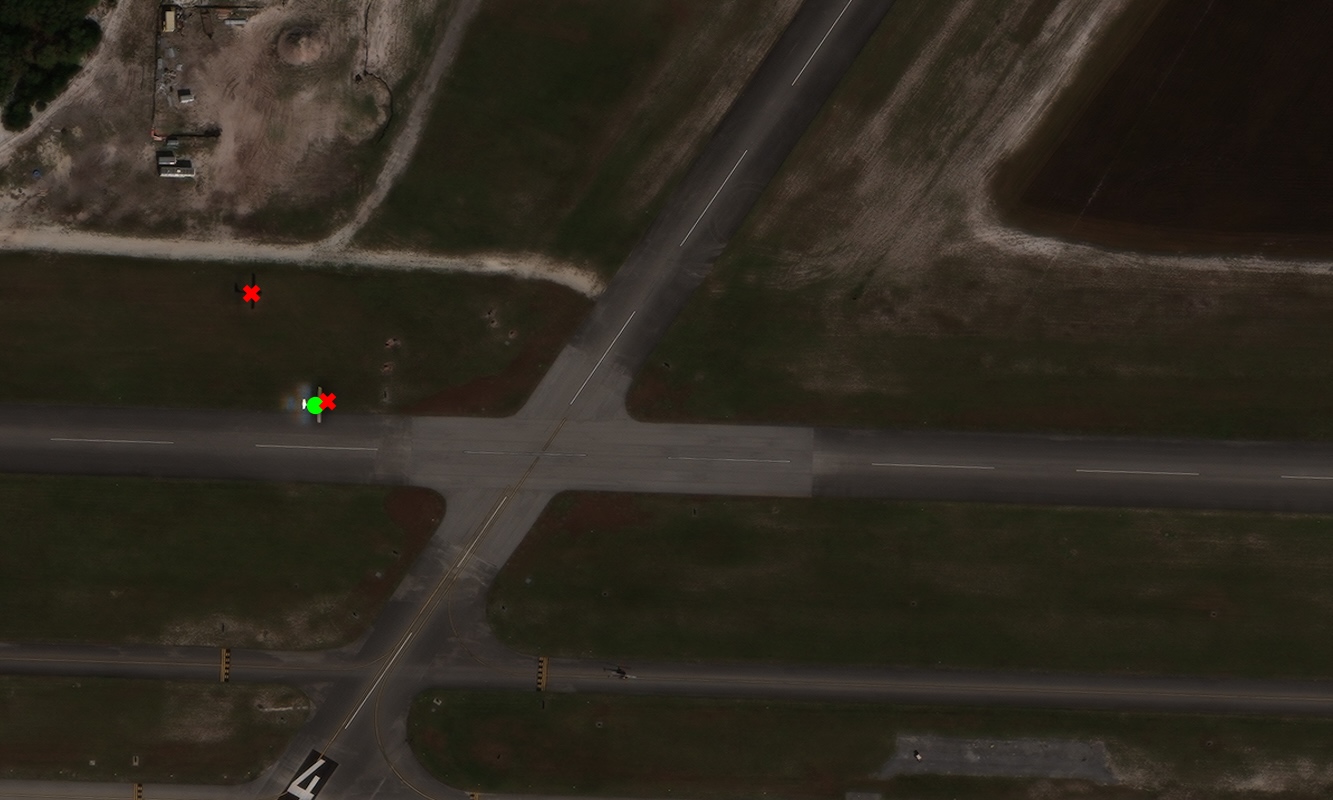}
    \caption{Illustration of a failure case on RarePlanes using Molmo 72B (ground truth represented by green dots and predictions represented by red Xs). An example of a failure where the model detects the plane's shadow as an additional plane. We noticed models making this mistake in other datasets as well.}
    \label{fig:rp_failure_2}
\end{figure*}

\begin{figure*}[h]
    \centering
    \includegraphics[width=\textwidth]{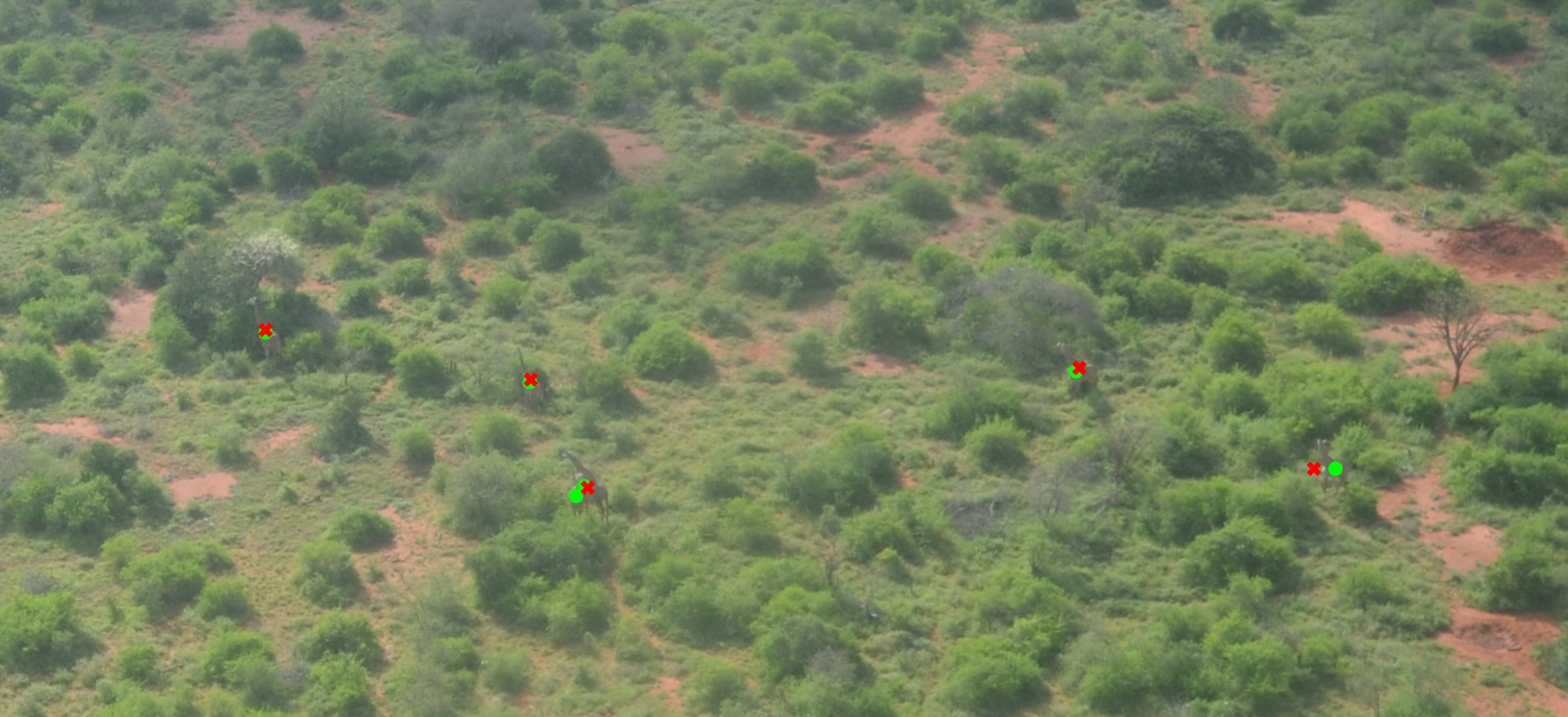}
    \caption{Zoomed in illustration of a success case on the Animal Population dataset using Qwen 2.5-VL 7B (ground truth represented by green dots and predictions represented by red Xs). Note how difficult it is to distinguish the giraffes from the background, nevertheless, Qwen successfully locates each one.}
    \label{fig:qwen_success}
\end{figure*}

\begin{figure*}[h]
    \centering
    \includegraphics[width=\textwidth]{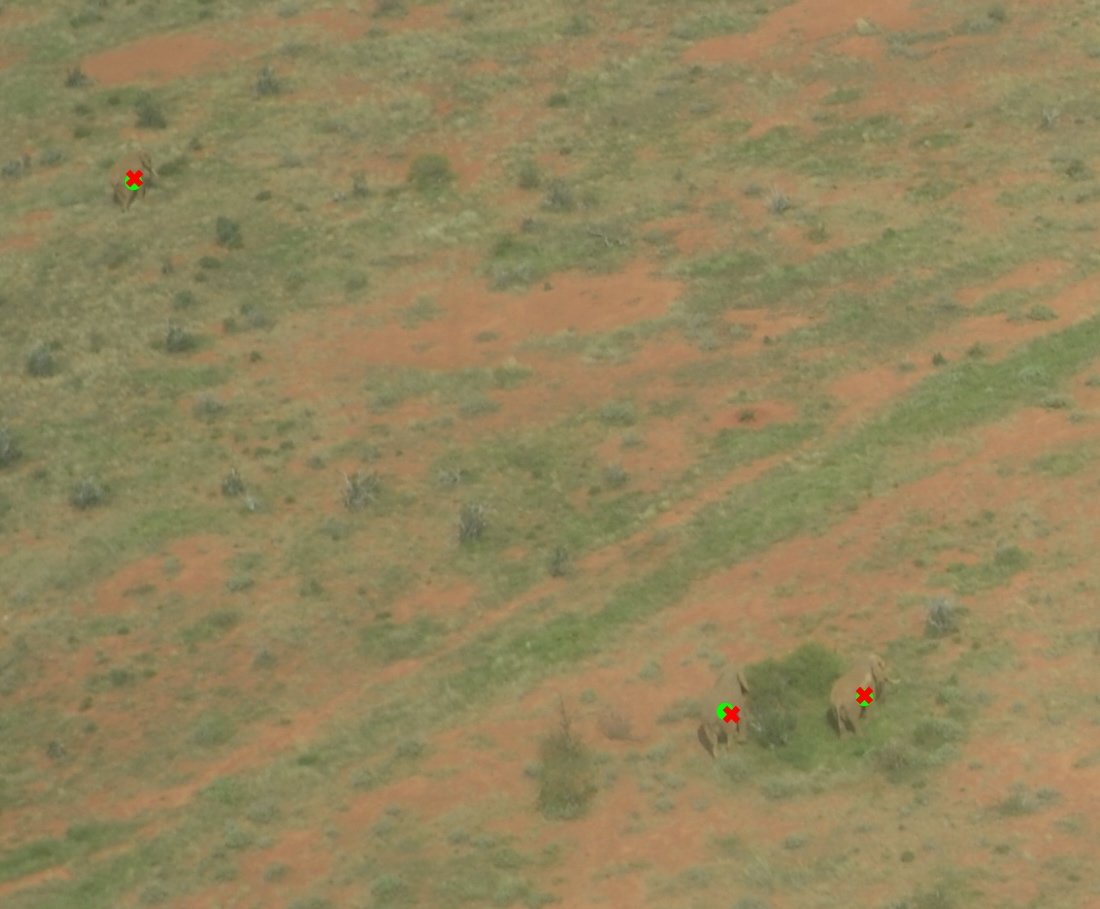}
    \caption{Zoomed in illustration of a success case on the Animal Population dataset using Qwen 2.5-VL 7B (ground truth represented by green dots and predictions represented by red Xs). Qwen successfully locates the elephants in the image. While they initially appear distinct, Figure \ref{fig:qwen_success_3} contains the zoomed out version of the image, where the elephants are no longer clear and are actually quite small relative to the scene.}
    \label{fig:qwen_success_2}
\end{figure*}

\begin{figure*}[h]
    \centering
    \includegraphics[width=\textwidth]{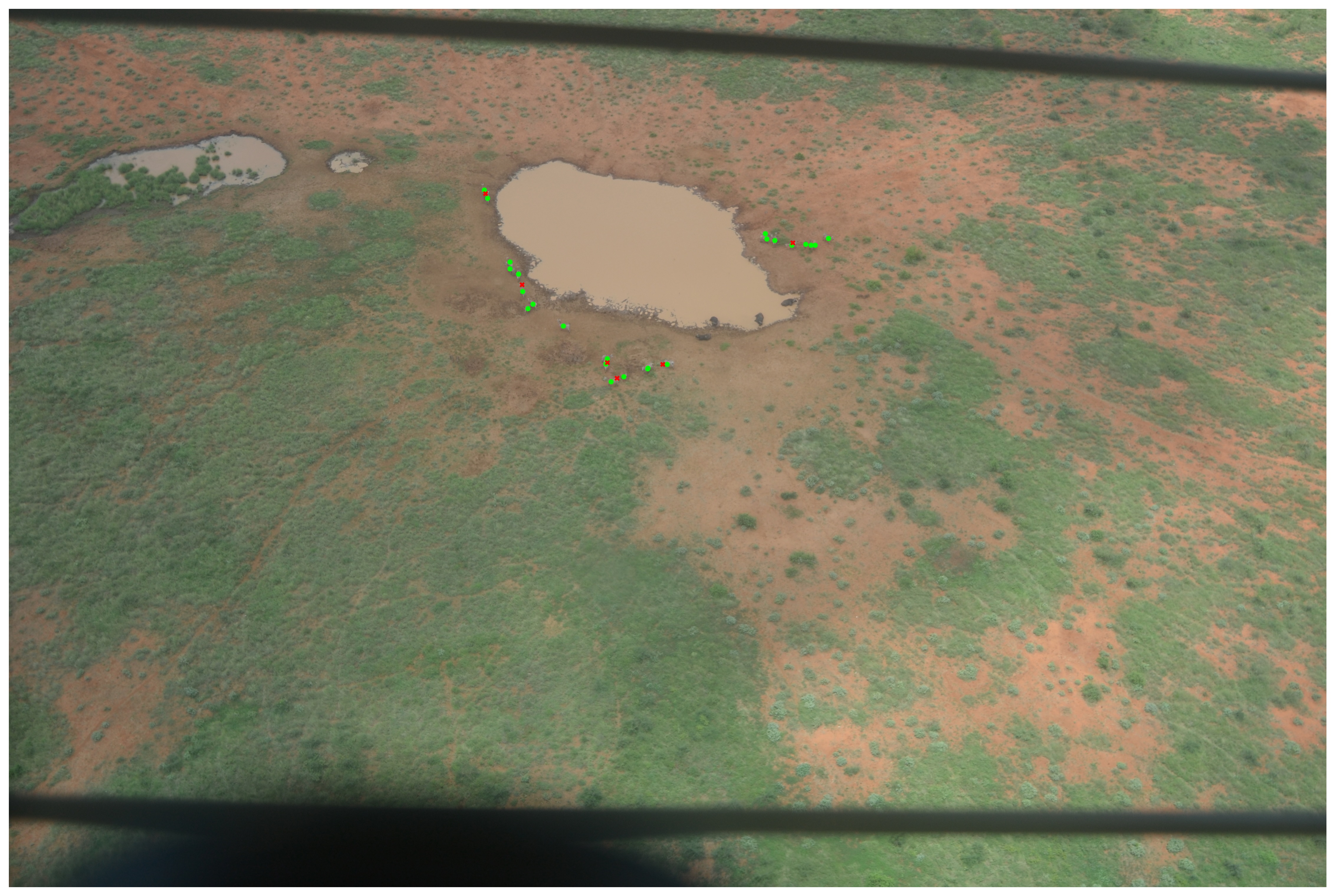}
    \caption{Full image from the Animal Population dataset with Qwen 2.5-VL 7B labels (ground truth represented by green dots and predictions represented by red Xs).}
    \label{fig:qwen_success_3}
\end{figure*}

\begin{figure*}[h]
    \centering
    \includegraphics[width=\textwidth]{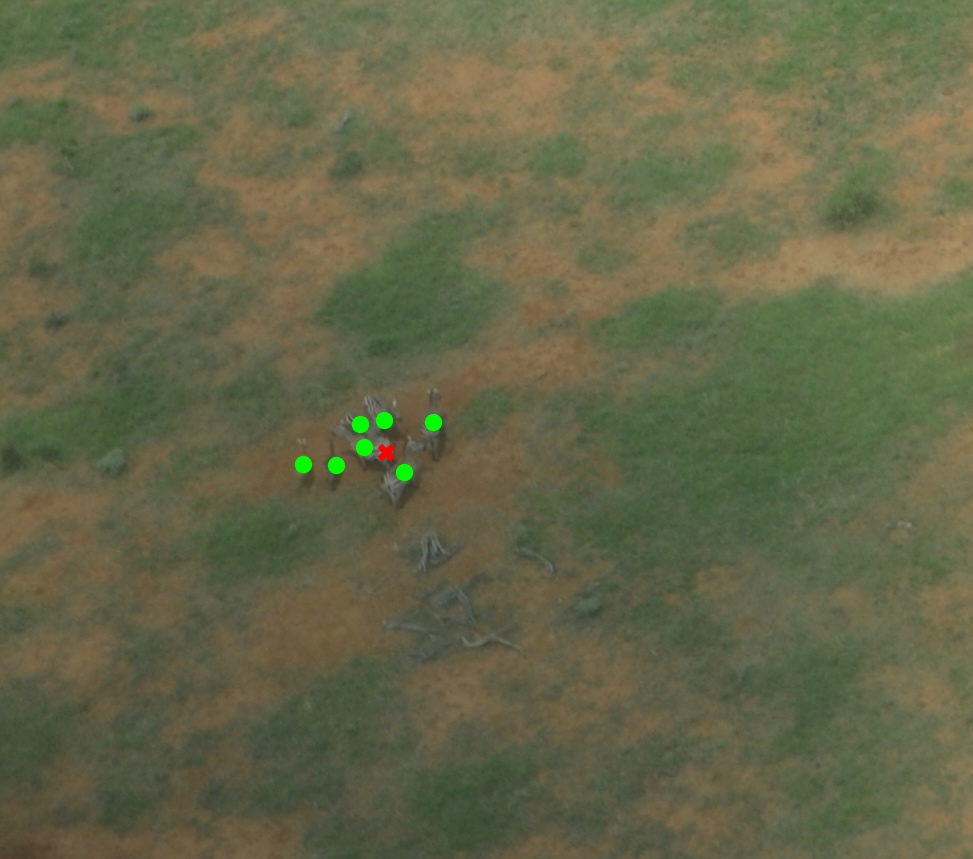}
    \caption{Zoomed in image from the Animal Population dataset with Qwen 2.5-VL 7B labels (ground truth represented by green dots and predictions represented by red Xs). This illustrates a common failure case where Qwen places a point on a group of animals, rather than individually identifying each one. This substantially hurts the AP of the model.}
    \label{fig:qwen_failure}
\end{figure*}

\begin{figure*}[h]
    \centering
    \includegraphics[width=\textwidth]{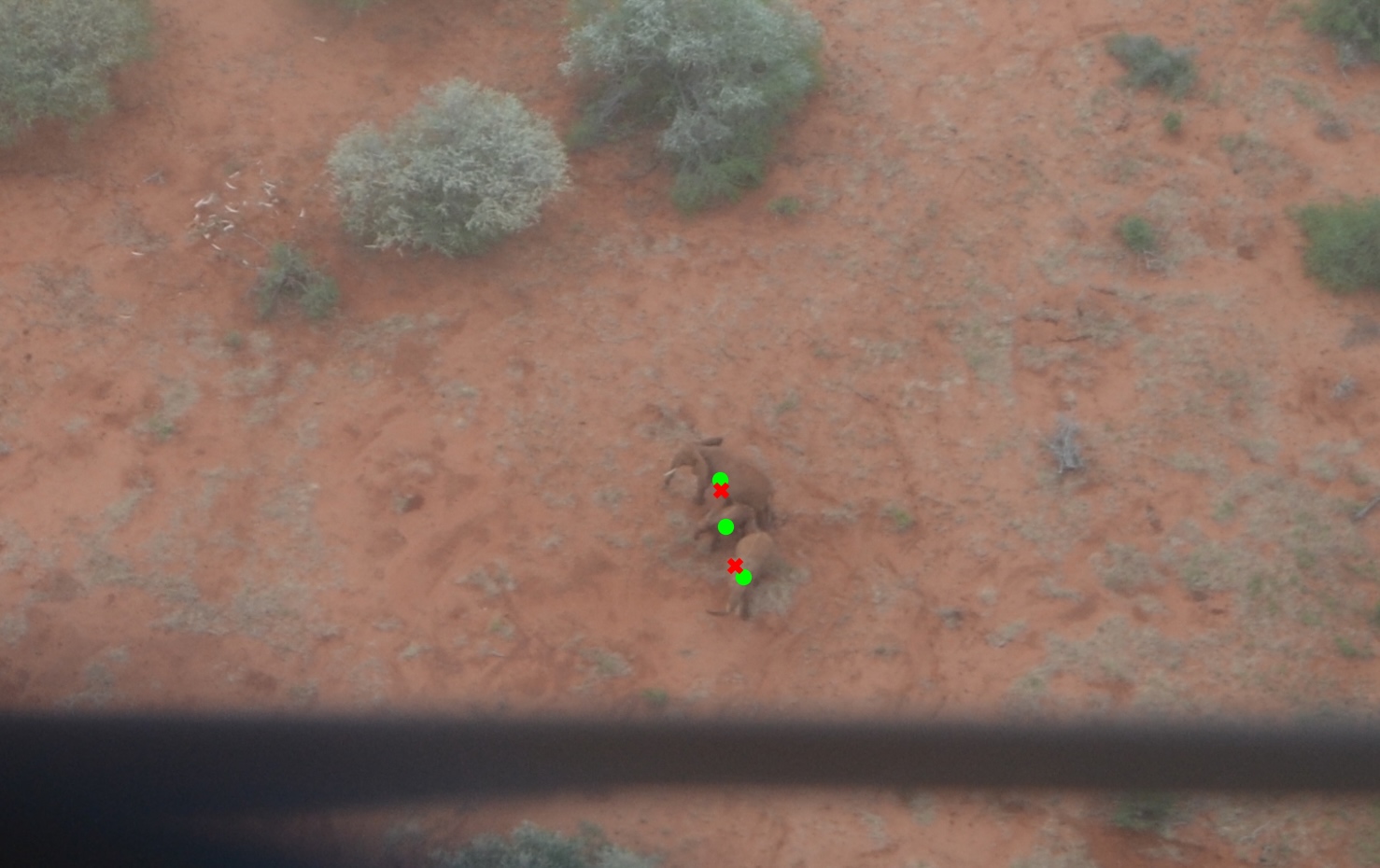}
    \caption{Zoomed in image from the Animal Population dataset with Qwen 2.5-VL 7B labels (ground truth represented by green dots and predictions represented by red Xs). This is a less severe failure case, more aligned with the types of mistakes one might expect to see, where one individual in a group of animals is missed.}
    \label{fig:qwen_failure_2}
\end{figure*}

\begin{figure*}[h]
    \centering
    \includegraphics[width=\textwidth]{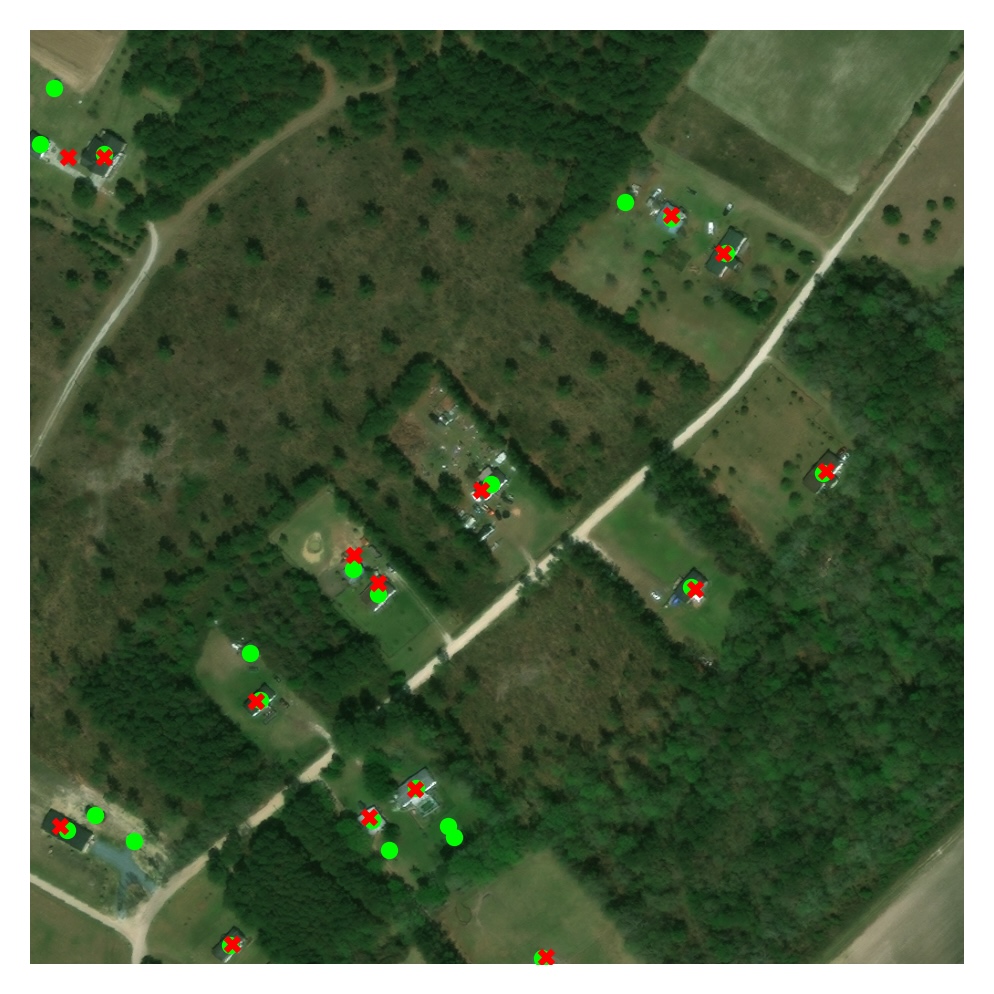}
    \caption{Image from xBD dataset with Molmo 72B labels (ground truth represented by green dots and predictions represented by red Xs). This is an example of a scenario where the model was relatively successful. It definitely possesses the required knowledge to detect buildings from an overhead angle. However, it misses certain buildings, especially when they are quite small, with some of them being fully subsumed by the small dots that we placed on the image.}
    \label{fig:xview_success}
\end{figure*}

\begin{figure*}[h]
    \centering
    \includegraphics[width=\textwidth]{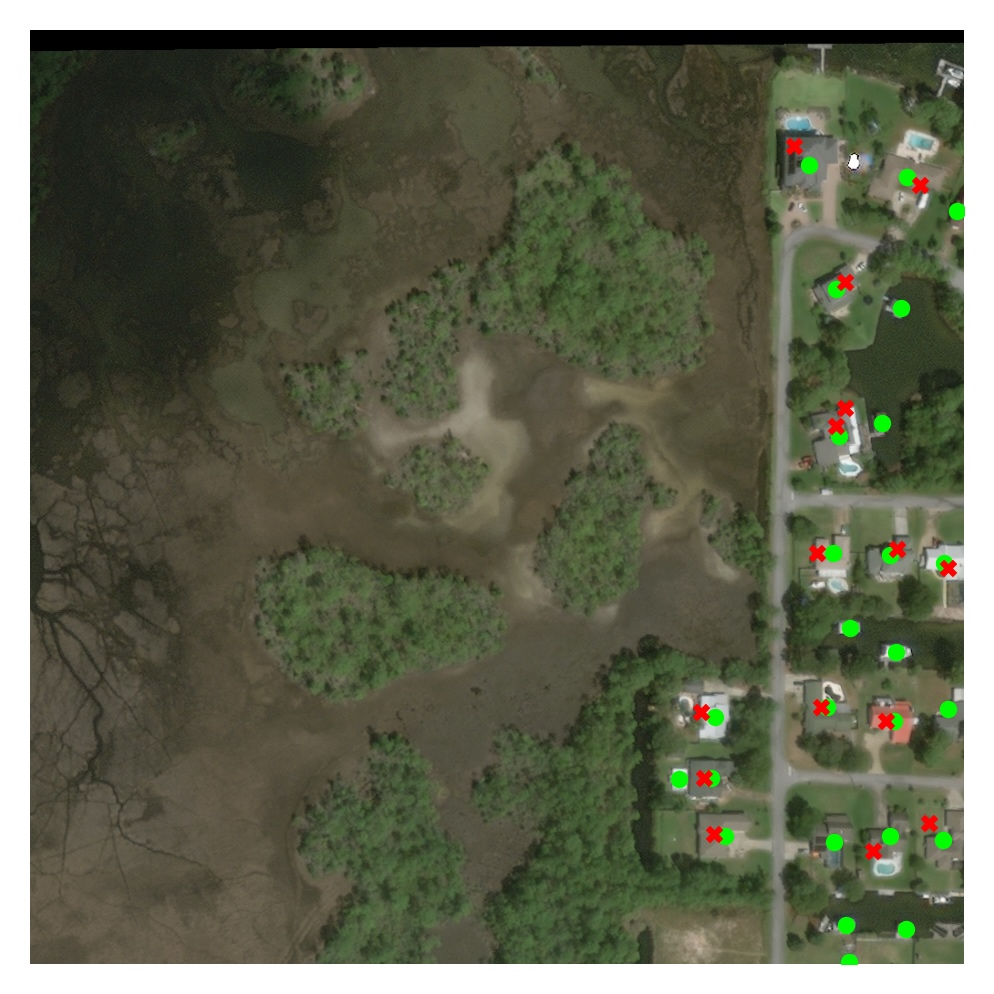}
    \caption{Image from xBD dataset with Molmo 72B labels (ground truth represented by green dots and predictions represented by red Xs). Another scene in which Molmo was relatively successful. In this case, it again detected many of the homes, which are already relatively small, mostly missing smaller buildings which appear to be sheds or garages.}
    \label{fig:xview_success_2}
\end{figure*}

\begin{figure*}[h]
    \centering
    \includegraphics[width=\textwidth]{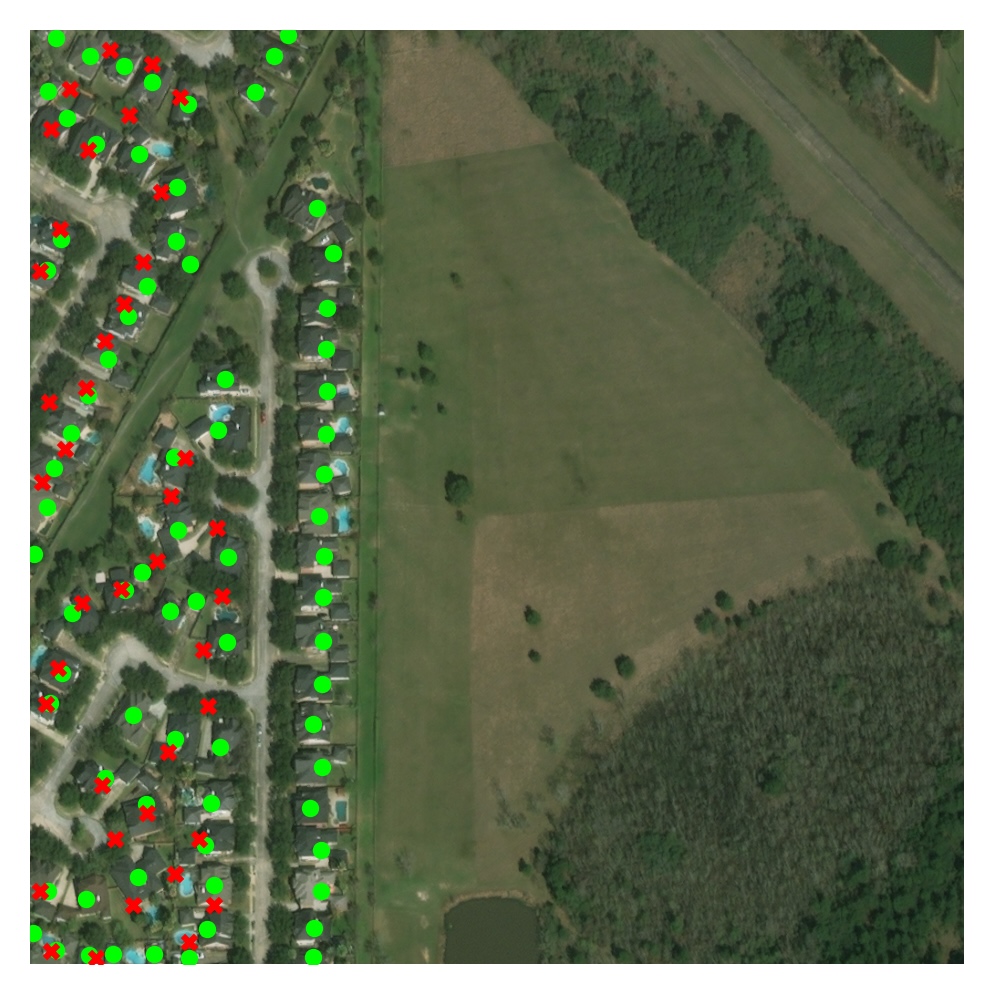}
    \caption{Image from xBD dataset with Molmo 72B labels (ground truth represented by green dots and predictions represented by red Xs). An exmaple of a failure scenario, where there are too many houses in the image for molmo to detect. It is worth noting that it does get many of the buildings. However, we found that even with expanding the generated token limit, Molmo falls apart past a certain number of objects in a given image, likely due to the distribution of object counts that it was trained on.}
    \label{fig:xview_failure}
\end{figure*}

\begin{figure*}[h]
    \centering
    \includegraphics[width=\textwidth]{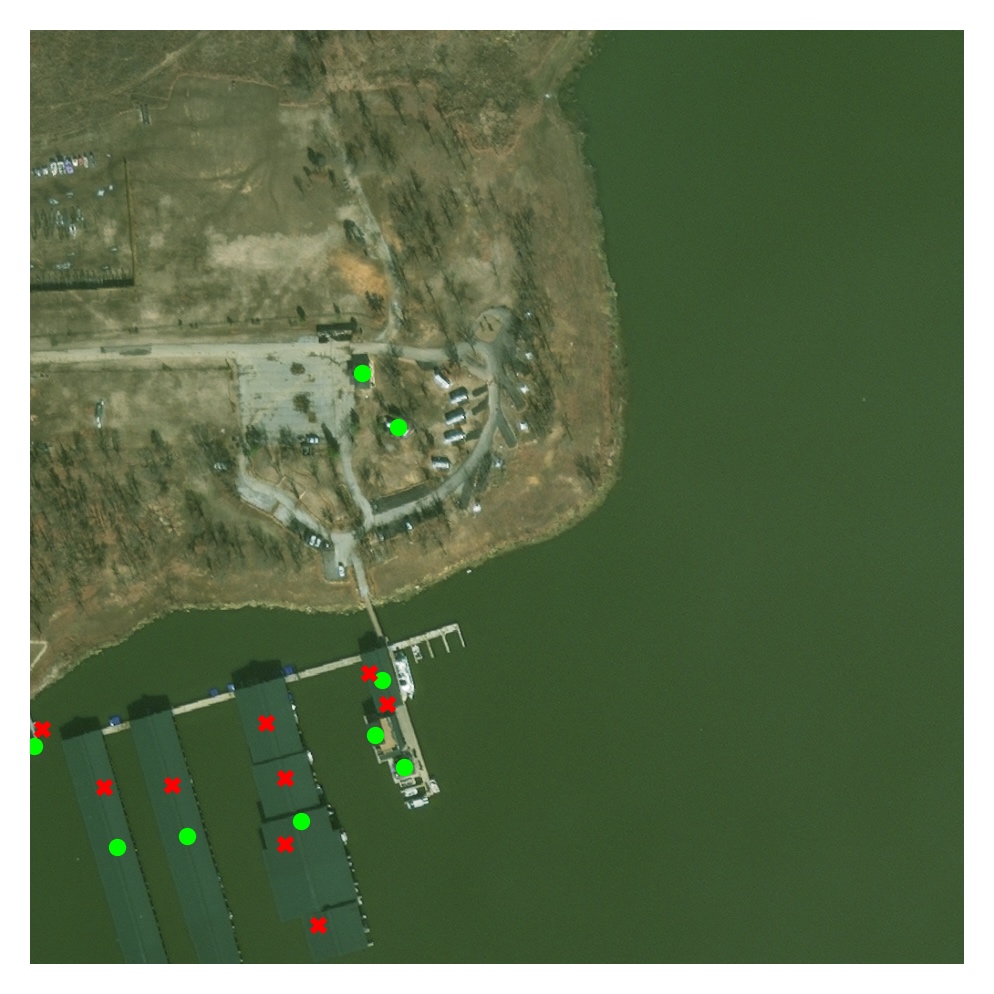}
    \caption{Image from xBD dataset with Molmo 72B labels (ground truth represented by green dots and predictions represented by red Xs). Another example of a failure scenario, where in this case the model incorrectly segments a larger building into multiple smaller buildings.}
    \label{fig:xview_failure_2}
\end{figure*}

\end{document}